\documentclass[conference]{IEEEtran}
\usepackage{color}
\usepackage{graphicx}
\usepackage[numbers]{natbib}
\usepackage{caption}
\usepackage{subcaption}
\usepackage{hyperref}
\usepackage{bookmark}
\usepackage{amsmath}
\usepackage[utf8]{inputenc}
\usepackage[stable]{footmisc}

\ifCLASSINFOpdf
\else
\fi

\begin{document}
%
\title{An empirical study on the effects of different types of noise in image classification tasks}

\newif\iffinal
\finaltrue
\newcommand{\jemsid}{98}

\iffinal
    \author{%
        \IEEEauthorblockN{Gabriel B. Paranhos da Costa, Welinton A. Contato, Tiago S. Nazare, João E. S. Batista Neto, Moacir Ponti}
        \IEEEauthorblockA{%
          Instituto de Ciências Matemáticas e de Computação (ICMC) -- Universidade de São Paulo (USP)\\
          São Carlos/SP -- 13566-590, Brazil\\
          Email: \{gbpcosta, welintonandrey, tiagosn\}@usp.br, jbatista@icmc.usp.br, ponti@usp.br} 
    }
\else
  \author{WVC paper ID: \jemsid \\ }
\fi


\maketitle

\begin{abstract}
Image classification is one of the main research problems in computer vision and machine learning. Since in most real-world image classification applications there is no control over how the images are captured, it is necessary to consider the possibility that these images might be affected by noise (e.g. sensor noise in a low-quality surveillance camera). In this paper we analyse the impact of three different types of noise on descriptors extracted by two widely used feature extraction methods (LBP and HOG) and how denoising the images can help to mitigate this problem. We carry out experiments on two different datasets and consider several types of noise, noise levels, and denoising methods. Our results show that noise can hinder classification performance considerably and make classes harder to separate. Although denoising methods were not able to reach the same performance of the noise-free scenario, they improved classification results for noisy data.
\end{abstract}


%

\section{Introduction} \label{sec:intro}

The study of noise in visual data is a matter of major interest within the image processing and computer vision communities. Due to that many different denoising algorithms were developed for both image~\cite{Dabov09} and video~\cite{Contato16} restoration. These methods are able to improve image quality in applications raging from microscopy~\cite{ponti2016image} to astronomy~\cite{beckouche2013astronomical} 


Over the last decades the image classification task has motivated the development of many image descriptors (e.g. LBP~\cite{Ojala94}, HOG~\cite{Dalal2005}) and, more recently, representation learning techniques~\cite{Bengio2013}. 
Nonetheless, the preprocessing stages of the image classification pipeline -- that could incorporate and benefit from denoising techniques -- have been neglected as pointed out by~\cite{Ponti16quantization, ponti2013compact,kanan2012color}. Moreover, little has been done to measure the impacts of different types of noise in image classification~\cite{Dodge16}, which can hinder the deployment of computer vision systems in scenarios where image quality varies.

Considering the above-mentioned gaps, in this paper we experimentally measure the effects of different types of noise on image classification and investigate denoising algorithms help to mitigate this problem. By doing so, we analyse our results based on the following topics: 
\begin{enumerate}
    \item Is the performance of a classifier hampered by noise when using the LBP and HOG methods to describe the image dataset?
    \item The decrease in performance is due to the fact that noise makes it harder to separate the classes or does the model learned from images without noise is not robust enough to deal with noisy images?
    \item Can denoising methods help in these situations?
\end{enumerate}

Our results show that classifiers suffer to generalise to different noisy data and image classification becomes harder when dealing with noisy images. Though denoising algorithms can help to mitigate the effects of noise, they may also remove important information, reducing classification performance.

\section{Related work}
\label{sec:relatedwork}

\begin{figure}[!ht]
    \centering
    \includegraphics[width=0.48\textwidth]{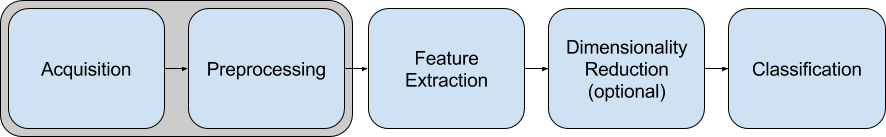}
    \caption{Classification pipeline. Our study focuses on the first two stages (highlighted by the gray box). This image was based on Figure 1 of~\cite{Ponti16quantization}.}
    \label{fig:classification_pipeline}
\end{figure}

\citet{Ponti16quantization} divide image classification in five stages (see Figure~\ref{fig:classification_pipeline}) and show that the method used to convert the images from RGB to grayscale can have a substantial impact on classification performance. They also demonstrate that RGB to grayscale conversion can be used as an effective dimensionality reduction procedure. Their results show that early stages of the classification pipeline -- despite being neglected in most image classification applications -- can directly influence classification performance. Some other papers~\cite{kanan2012color,ponti2013compact} also point out the importance of these early stages. Nonetheless, as in~\cite{Ponti16quantization}, they only focus on RGB to grayscale conversion and do not consider noisy images.%

\citet{Dodge16} analyse how image quality can hamper the performance of some state-of-the-art deep learning models by using networks trained on noise-free images to classify noisy, blurred and compressed images. Their results show that image classification is directly affected by image quality. Similarly, \citet{Kylberg13} evaluate noise robustness of several LBP variants. Given that on both these papers the classifiers are trained in noise-free images, it is not possible to infer if the learned models are not able to deal with noisy images or if noise makes the classes harder to separate.

\section{Technical background}
\label{sec:tecback}


\subsection{Local binary patterns}

Local Binary Patterns (LBP)~\cite{Ojala94} is a texture-based image descriptor that, due to its success, has several variants and improved versions~\cite{Nanni12, Ojala02}. In this paper, we employ the version that uses uniform patterns and it is invariant to gray scale shifts, scale changes and rotations. This variant achieves good results while generating low dimensional features.

The \emph{LBP descriptor} is the distribution (a histogram) of texture patterns extracted for every pixel in an image. Thus, prior to computing the LBP descriptor, it is necessary to compute a texture pattern representation for each pixel. Such texture representation is called \emph{LBP code} and it is based on the difference between a pixel and its neighbors. These neighbors can be arranged in a circle or in a square. A neighborhood is defined by the parameters $R$ and $P$, where $P$ is the number of neighbors and $R$ is the radius of the circular neighborhood (or the side of the square neighborhood). If one of the neighbors is not at the center of a pixel, its value needs to be obtained via interpolation. 

The LBP code (a binary code) for a pixel $g_c$ and its neighbors is defined as follows:
\begin{equation}\label{lbp_pr}
    LBP_{P,R} = \sum_{p=0}^{P-1} s(g_p - g_c)2^p,
\end{equation}
where $s$ is the sign function and $g_0, \dots, g_{P-1}$ are neighbors of $g_c$. This LBP code is invariant to grayscale shifts, because it is based on the differences of pixels and not in absolute values. Also, since only the sign of the difference result is considered, the code is invariant to scale. On the other hand, such code is not invariant to rotation. 

It is possible to achieve some invariance to rotation by using the following LBP code: 
\begin{equation}
    LBP_{P,R}^{ri} = \min\{ROR(LBP_{P,R}, i)~|~i = 0,...,P-1\},
\end{equation}
where $ROR(c, i)$ is the result of $i$ circular right bit-wise shifts applied to the code $c$. As an example, if $c = 01110010$ and $i = 2$, then $ROR(c,i) = \mathbf{10}011100$. By always considering the minimum of all possible bit-wise shifts, a code that is more robust to rotations can be obtained.

\citet{Pietikainen00} discovered that when LBP patterns are considered circularly, they usually contain two or less bit transitions (patterns with such characteristic were named uniform). The other patterns -- that have more than two transactions -- occur rarely and were called non-uniform.

Finally, to obtain the LBP descriptor -- up to now we were talking about LBP codes -- a histogram is computed. In this histogram, each uniform pattern has its own bin, while there is one bin for all the non-uniform patterns. 

\subsection{Histogram of oriented gradients}

Based on evaluating well-normalized local histograms of image gradient orientations in a dense grid, Histogram of Oriented Gradients (HOG)~\cite{Dalal2005} takes advantage of the distribution of local intensity gradients or edges directions to characterize the local object appearance and shape. This is done by diving the image window into small connected regions, called cells, in which a local histogram of gradient directions or edge directions is computed over all pixels. The final representation is obtained by combining the histograms computed in all cells of the image. HOG descriptors are particularly suited for human detection~\cite{Dalal2005}.

To extract HOG descriptors from an image, firstly, gradient values must be computed. This is most commonly done by filtering the color or intensity data of the image using the one-dimensional centered point discrete derivative mask in the horizontal ($[-1, 0, 1]$) and vertical directions ($[-1, 0, 1]^{T}$). Then, the image is divided into small cells of rectangular (R-HOG) or circular shape (C-HOG). Each pixel contained by a cell is used in a weighted manner to create a orientation-based histogram. This histogram is created for each cell and its bins are evenly spread over the orientation of the gradients. The range of the orientation can be defined over 0 to 180 degrees or over 0 to 360 degrees, depending if the gradient is ``signed'' or ``unsigned''. The contribution of a pixel to each bin of the histogram is weighted based on the magnitude of the gradient or some function of this magnitude.

To increase robustness to illumination and contrast changes, gradient strengths are locally normalized by grouping cells together into blocks. Some methods commonly used for normalization are: $\ell_{2}$-norm (Equation~\ref{eq:l2norm}), hysteresis-based $\ell_{2}$ normalization~\cite{Lowe2004} or $\ell_{1}$-sqrt (Equation~\ref{eq:l1sqrt}), where $\nu$ is the non-normalized vector containing all histograms of a given block, $\left\lVert \Delta \nu \right\rVert_{k}$ is its $k$ norm for $k = 1, 2$ and $e$ is a small constant. 

\begin{equation}
\label{eq:l2norm}
    f = \frac{\nu}{ \sqrt{ \strut \lVert \Delta \nu \rVert^{2}_{2} + e^{2} } }
\end{equation}

\begin{equation}
\label{eq:l1sqrt}
    f = \sqrt{ \strut \frac{\nu}{ (\lVert \Delta \nu \rVert_{1} + e) } }
\end{equation}

Blocks typically overlap, which means that a cell can contribute to more than one block, and, therefore, to the final descriptor. The size and shape of the cells and blocks and the number of bins in each histograms are set by the user.

\subsection{Median filter}

The Median filter replaces each pixel value by the median pixel value in a $n \times n$ neighborhood centered on it. This filter can be described by the following equation:
\begin{equation}
   \hat{z}(x,y) = median(z_k \mid k=1,...,n\times n),
\label{eq:median}
\end{equation}
where $z_k$ for $k=1,...,n\times n$ are the pixel values within the neighborhood centered on $(x,y)$.

\subsection{Non-Local Means}

The Non Local Means (NLM) originally presented in~\cite{buades2005non} has inspired several variations. In this paper we use the windowed version as proposed by~\citet{Buades11}. Given a noisy image $v$, this NLM variant defines a restored version pixel $i$ as a weighted average of all pixels inside of a window of size $s \times s$ centered on $i$ using the following equation:
\begin{equation}\label{eq_nlm_pixel}
	NL[v](i) = \sum_{j \in S_i} w(i,j)v(j),
\end{equation}
where the weight $w(i,j)$ measures the similarity between pixels $i$ and $j$ and $S_i$ is the $s \times s$ search window ($s$ is an user-defined parameter). Each $w(i,j)$ is computed as follows:
\begin{equation} \label{eq_w}
	w(i,j)=\frac{1}{Z(i)} e^{-\frac{{||v(\mathcal{N}_i)-v(\mathcal{N}_j)||}^2_{2,a}}{h^2}},
\end{equation}
where $\mathcal{N}_i$ and $\mathcal{N}_j$ are $p \times p$ regions centered at $i$ and $j$ ($p$ is an user-defined parameter) and $h$ is an user-defined parameter that represents filtering level. To compute the similarity between $\mathcal{N}_i$ and $\mathcal{N}_j$ an Euclidean distance weighted by a Gaussian kernel with standard deviation defined by the user-defined parameter $a$ is used. 

\section{Experiments}
\label{sec:methodology}

\subsection{Experimental setup}

To evaluate if noise hampers classification performance we generated noisy versions of two datasets (Corel and Caltech101-600) using different levels of three types of noise: Gaussian, Poisson and salt \& pepper. Moreover, to understand the impacts of employing a denoising algorithm as preprocessing, we restored these noisy images using two denoising methods: Median filter and Non-Local Means. All these operations were performed on both, training and test, sets of both datasets. 

We trained different linear Support Vector Machines (SVMs) for every version of their training set. Given that every training set version only has one type of noise (or no noise at all), a model specialized on each level of each type of noise was created. 

\begin{figure}[ht]
    \centering
    \includegraphics[width=0.4\textwidth]{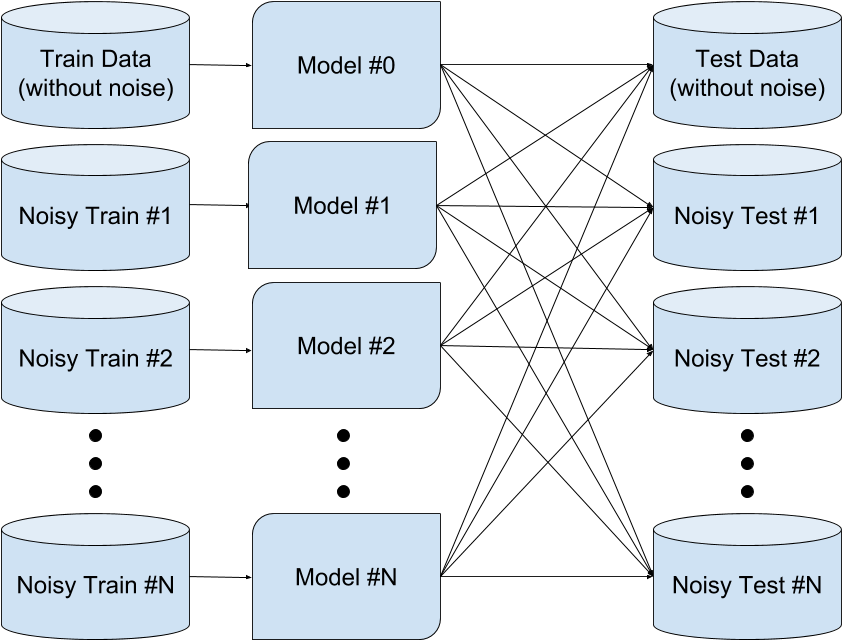}
    \caption{Experimental setup diagram. A different model is trained for every noisy version of the training set. Then, these models are evaluated on all versions of the test set.} \label{fig:diag_metod}
\end{figure}

Then, these models were used to classify every version of the test set. As with the training sets, each test set version also only contains one type of noise (or no noise at all), this allows the experiments to measure how well a model learned on a particular noisy training set performs on other types of noisy images (see Figure~\ref{fig:diag_metod} for a diagram that summarizes this setup). In addition, by training a model using a certain type of noise and noise level and then evaluating its performance on a test set with the same characteristics, it is possible to make a superficial analysis on the linear separability of the problem (since linear SVMs were used). 


Since the selected datasets have more than two classes, we trained SVM models using a ``one-vs-all'' approach. Furthermore, to evaluate their performance, an average F1-score weighted by the number of instances in each class was used. This performance measure was chosen because it addresses the problem of evaluating the classification of unbalanced domains, that is, when classes have different number of instances. 

\subsection{Datasets}
\label{sec:dataset}

\subsubsection{Corel\texorpdfstring{\footnote{The Corel dataset is available at: \url{https://sites.google.com/site/dctresearch/Home/content-based-image-retrieval}}}{}} 

a dataset containing $10800$ RGB images of 80 classes, where each class has at least 100 images. Sample images from this dataset are shown in the first row of Figure~\ref{fig:Caltech101-600}.

   

\begin{figure}[!ht]
    \centering
    \begin{subfigure}[b]{0.105\textwidth}
            \includegraphics[width=\textwidth]{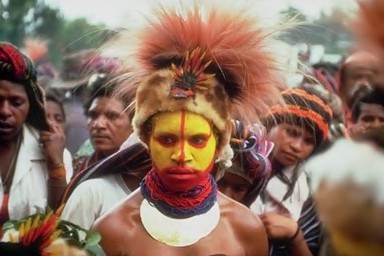}
    \end{subfigure}\quad
    \begin{subfigure}[b]{0.105\textwidth}
            \includegraphics[width=\textwidth]{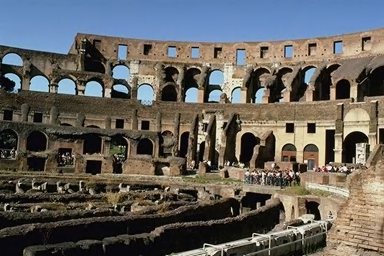}
    \end{subfigure}\quad
    \begin{subfigure}[b]{0.105\textwidth}
            \includegraphics[width=\textwidth]{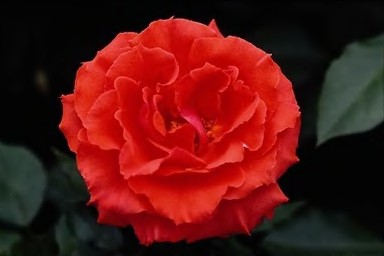}
    \end{subfigure}\quad
    \begin{subfigure}[b]{0.105\textwidth}
            \includegraphics[width=\textwidth]{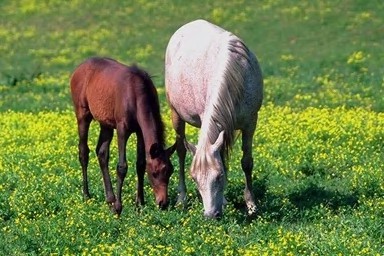}
    \end{subfigure}
    \\~\\
    \begin{subfigure}[b]{0.105\textwidth}
            \includegraphics[width=\textwidth]{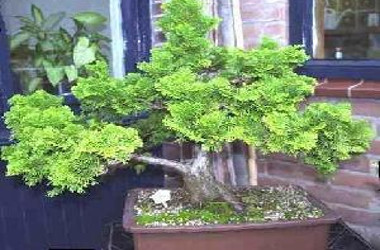}
    \end{subfigure}\quad
    \begin{subfigure}[b]{0.105\textwidth}
            \includegraphics[width=\textwidth]{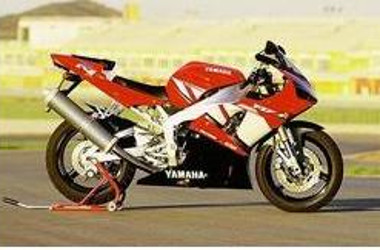}
    \end{subfigure}\quad
    \begin{subfigure}[b]{0.105\textwidth}
            \includegraphics[width=\textwidth]{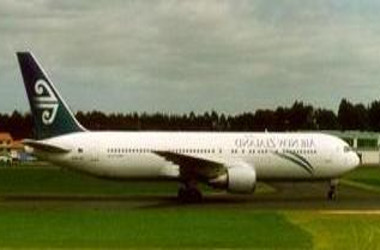}
    \end{subfigure}\quad
    \begin{subfigure}[b]{0.105\textwidth}
            \includegraphics[width=\textwidth]{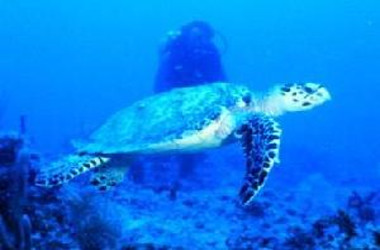}
    \end{subfigure}
   
    \caption{Sample images from the Corel (first row) and Caltech101-600 (second row) dataset~\cite{Ponti16quantization}.}
    \label{fig:Caltech101-600}
\end{figure}

\subsubsection{Caltech101-600\texorpdfstring{\footnote{The Caltech101-600 dataset is available at: \url{http://www.icmc.usp.br/pessoas/moacir/data/}}}{}} 

a subset~\cite{Ponti16quantization} of Caltech101~\cite{Fei2007} containing 6 classes (\emph{airplanes, bonsai, chandelier, hawksbill, motorbikes}, and \emph{watch}), each one with 100 examples. Images from this dataset can be seen in the second row of Figure~\ref{fig:Caltech101-600}.

\subsection{Reproducibility remarks and parameter values}

\begin{figure}[ht]
    \centering
    \begin{subfigure}[b]{0.14\textwidth}
            \includegraphics[width=\textwidth]{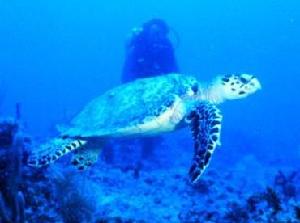}
            \caption{Original}
    \end{subfigure}\quad
    \begin{subfigure}[b]{0.14\textwidth}
            \includegraphics[width=\textwidth]{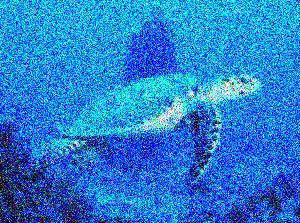}
            \caption{$\sigma = 20$}
    \end{subfigure}\quad
    \begin{subfigure}[b]{0.14\textwidth}
            \includegraphics[width=\textwidth]{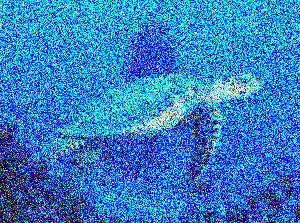}
            \caption{$\sigma = 50$}
    \end{subfigure}
   
    \caption{Example image from the Caltech101-600 dataset with different levels of Gaussian noise.}
    \label{fig:noise}
\end{figure}

Regarding the insertion of noise to the images, we considered three types of noise: Gaussian, Poisson and salt \& pepper. First, for the Gaussian noise, we used zero mean and five different values for the standard deviation ($\sigma = \{10, 20, 30, 40, 50\}$). Figure~\ref{fig:noise} shows an example of different levels of Gaussian noise. Secondly, since the Poisson is a noise dependent signal, to adjust the intensity of the Poisson noise applied to each image, it was necessary to multiply the image by a scale factor after generating the noise, controlling its effect on the image\footnote{An in depth explanation about the scale factor for the Poisson noise can be found at: \url{https://ruiminpan.wordpress.com/2016/03/10/the-curious-case-of-poisson-noise-and-matlab-imnoise-command/}}. In our tests we used five different levels for the Poisson scale factor (scale = $\{10, 10.5, 11, 11.5, 12\}$). Finally, the salt \& pepper noise was applied to each pixel with five different probabilities: $p = \{0.1, 0.2, 0.3, 0.4, 0.5\}$.

 
For the image descriptors, the parameters were fixed for all datasets and all noise types and levels. The LBP method was computed using radius $R = 1$ and circular neighborhood $P = 8$, while, for the HOG method, $8$ possible gradients orientations were considered and each cell was composed by a region of $16 \times 16$ pixels, while each block contains a single cell. To obtain fixed-size feature vectors using the HOG descriptor, all images in the dataset were resized. This process was carried out in three steps. First, considering the number of rows and columns, we computed the values of the bigger and the smaller dimension of every image. Second, we obtained the smallest value for the bigger and the smaller dimensions among all images within the dataset. Thirdly, given $b$ (the smallest value for the bigger dimension) and $s$ (the smallest value for the smaller dimension), we resized all images in a dataset so that they end up with their bigger dimension equals to $b$ and their smaller dimension equals to $s$. This procedure reduces the distortion caused by the resize.

\begin{figure}[ht]
    \centering
    \begin{subfigure}[b]{0.14\textwidth}
            \includegraphics[width=\textwidth]{images/ori.jpg}
            \caption{Original}
    \end{subfigure}\quad
    \begin{subfigure}[b]{0.14\textwidth}
            \includegraphics[width=\textwidth]{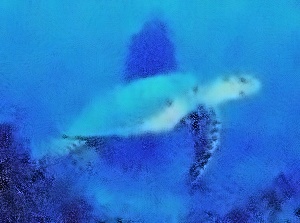}
            \caption{$\sigma = 20$}
    \end{subfigure}\quad
    \begin{subfigure}[b]{0.14\textwidth}
            \includegraphics[width=\textwidth]{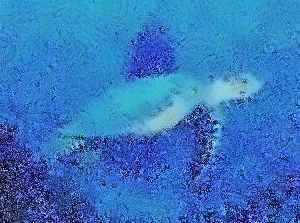}
            \caption{$\sigma = 50$}
    \end{subfigure}
   
    \caption{Images from the Caltech101-600 dataset with different Gaussian noise levels and after applying NLM.}
    \label{fig:NLM_gaussian}
\end{figure}

Concerning denoising methods, all NLM restored images were generate using $p = 7$, $s = 21$ (which are recommended by the original paper~\cite{buades2005non}) and $h = 25$. With the Median filter we used a neighborhood of $11 \times 11$ pixels. Examples of the images obtained after applying a denoising method can be seen in Figure~\ref{fig:NLM_gaussian}. During the classification stage, the parameters used to train each SVM were selected using a grid-search performed in a 5-fold cross validation on the training set. 

Due to reproducibility purposes, the code used in our experiments is available online\footnote{Repository url:
\iffinal
    \href{https://github.com/gbpcosta/wvc\_2016\_noise}{https://github.com/gbpcosta/wvc\_2016\_noise}
\else
    Due to anonymity purposes this url was omitted.
\fi
}.

\section{Results and discussion}
\label{sec:results}

\begin{figure}[ht]
    \centering
    \includegraphics[width=0.4\textwidth]{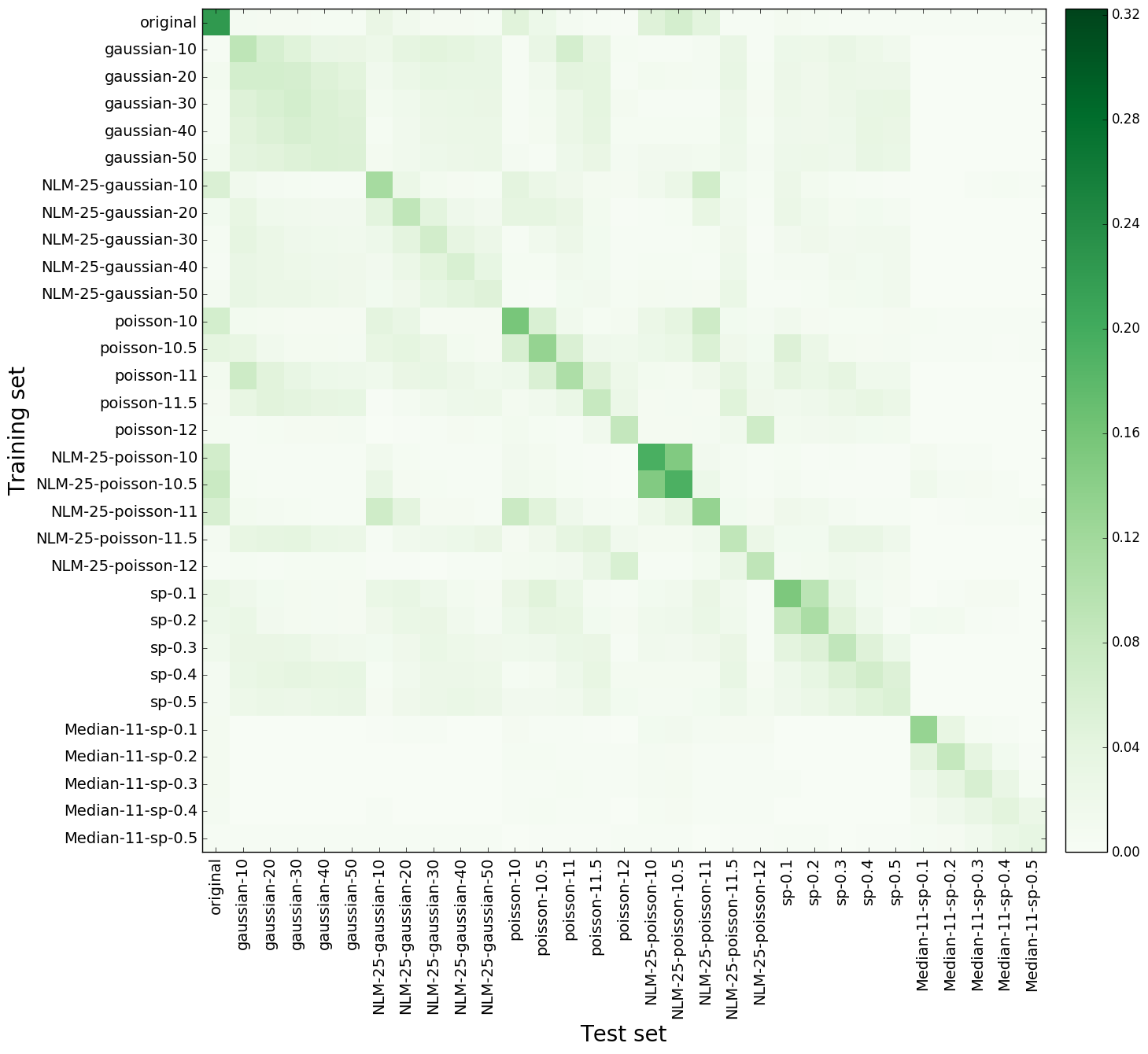}
    \caption{LBP results for the Corel dataset.}
    \label{fig:corel_lbp}
\end{figure}

\begin{figure}[ht]
    \centering
    \includegraphics[width=0.4\textwidth]{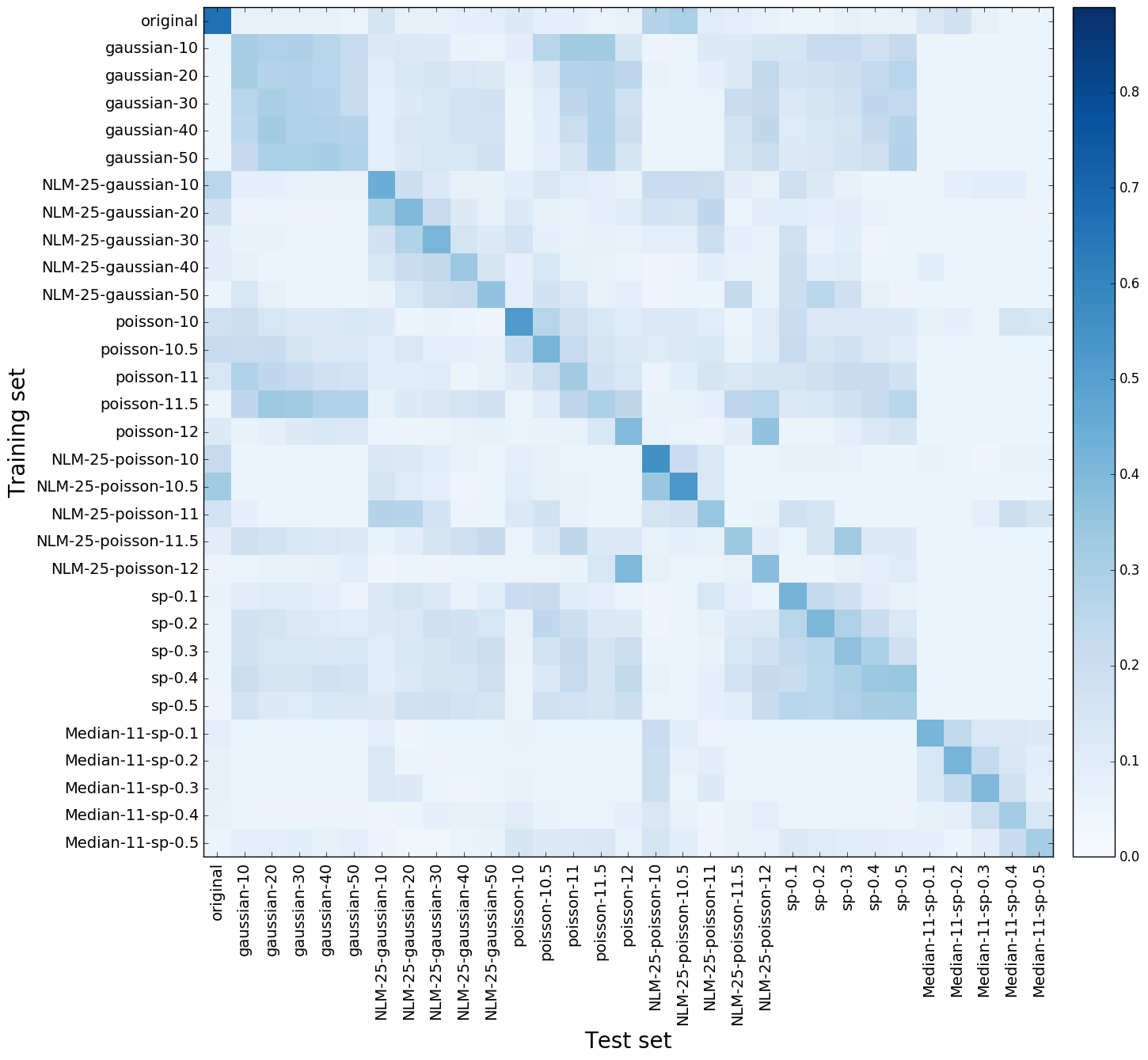}
    \caption{LBP results for the Caltech101-600 dataset.}
    \label{fig:caltech600_lbp}
\end{figure}

\begin{figure}[ht]
    \centering
    \includegraphics[width=0.4\textwidth]{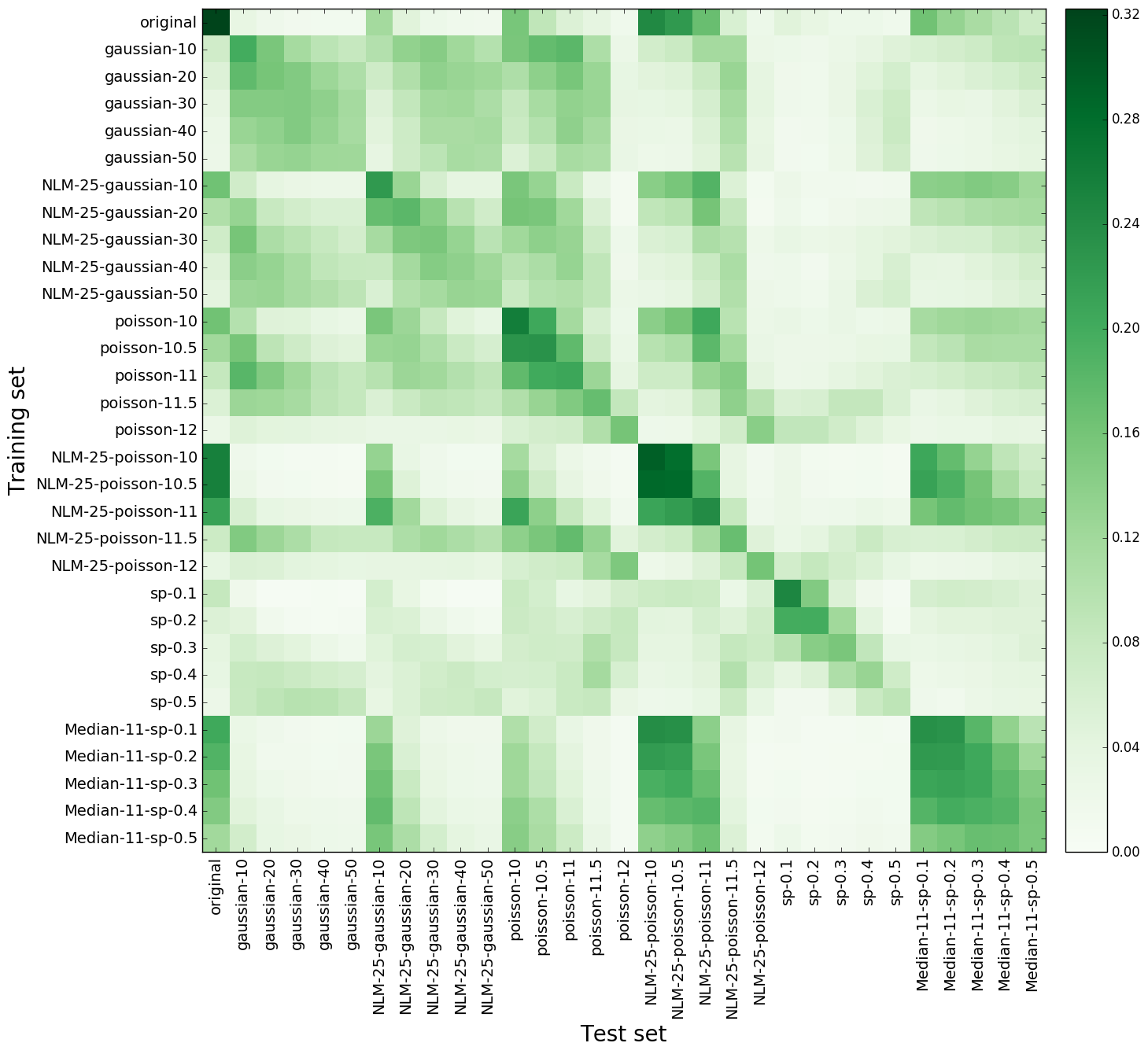}
    \caption{HOG results for the Corel dataset.}
    \label{fig:corel_hog}
\end{figure}

\begin{figure}[ht]
    \centering
    \includegraphics[width=0.4\textwidth]{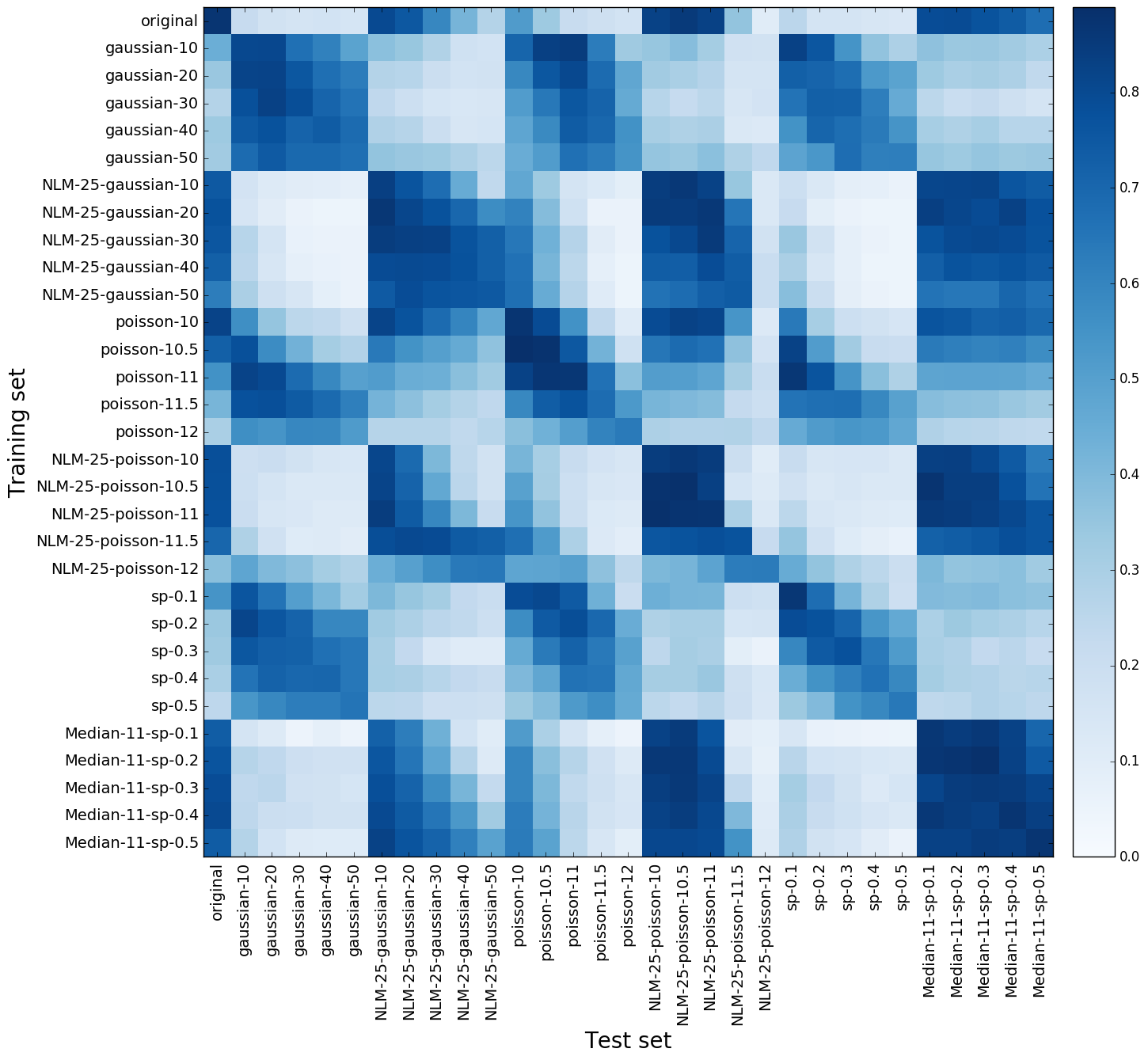}
    \caption{HOG results for the Caltech101-600 dataset.}
    \label{fig:caltech600_hog}
\end{figure}

Using the experimental setup presented earlier, in this section we analyse our results considering the three questions presented in Section~\ref{sec:intro}. As can be noticed by the heatmaps presented in Figures~\ref{fig:corel_lbp},~\ref{fig:caltech600_lbp},~\ref{fig:corel_hog} and~\ref{fig:caltech600_hog}, the HOG descriptor obtained better results in both datasets. However, our goal is not to compare the descriptors, but rather analyse the impact of noise in image classification by shedding some light on the following questions. 

~

\noindent\textbf{1) Is the performance of a classifier hampered when using the LBP and HOG methods to describe a noisy image dataset?} 

\begin{figure}[ht]
    \centering
    \includegraphics[width=0.45\textwidth]{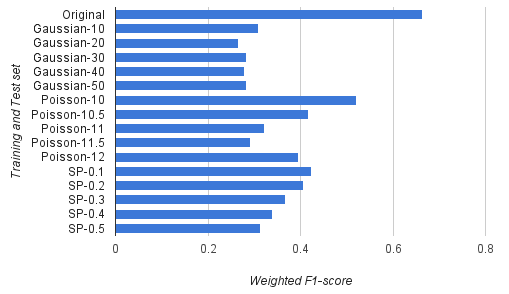}
    \caption{LBP results for the Caltech101-600 when both training and testing was performed with the same type and level of noise.}
    \label{fig:diagonal}
\end{figure}

To answer this question we created Figures~\ref{fig:corel_lbp},~\ref{fig:caltech600_lbp},~\ref{fig:corel_hog} and~\ref{fig:caltech600_hog}. Each one of these figures is a heatmap representing the F1-score levels obtained by a classifier in all versions of a dataset (noisy, original and restored). It is possible to observe that the best results were obtained by classifiers trained and tested using noise-free images. This means that, for the analysed scenarios, image classification using LBP and HOG descriptors classified by a linear SVM, is hampered when using noisy images as input. Additionally, the higher the noise level the lower the F1-score (see Figure~\ref{fig:diagonal}) if we consider a model trained with the original (noise-free) train data. This effect was observed also in previous studies~\cite{Kylberg13, Dodge16}, but for other descriptors, classifiers, datasets, and types of noise. 

Please notice that the darkest color in the heatmaps is defined by the best result obtained in that dataset during the experiments and \textbf{not} by $1.0$ (the best possible F1-score value). For that reason the scale of Figures~\ref{fig:corel_lbp} and~\ref{fig:corel_hog} is different from the one of Figures~\ref{fig:caltech600_lbp} and~\ref{fig:caltech600_hog}.

\noindent\textbf{2) The decrease in performance is due to the fact that noise makes it harder to separate the classes or does the model learned from images without noise is not robust enough to deal with noisy images?}

\begin{figure}[ht]
   \centering
    \includegraphics[width=0.48\textwidth]{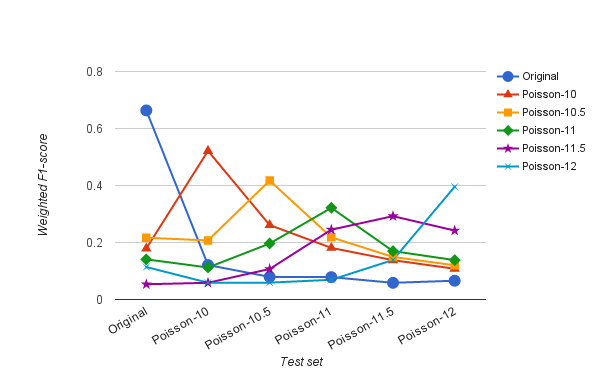}
    \caption{Comparison of the LBP results for the Corel dataset when training and testing is performed using images affected by the Poisson noise.}
    \label{fig:lbp-poisson-poisson}
\end{figure}

If we look at Figure~\ref{fig:lbp-poisson-poisson} it is possible to see that the models trained in a specific noise configuration have the best performance for a test set with the same noise configuration. Nevertheless, if we compare these best results for every noise level (as shown in Figure~\ref{fig:diagonal}), the best F1 -- for both descriptors in both datasets -- are obtained when both training and test is noise-free. Therefore, given that all these models where build after a grid search and that linear SVMs were used, our results indicate that the classes become less linearly separable in the presence of noise.

Those results show that LBP and HOG are sensitive to noise, which might cause it to produce different feature spaces for the same data under different levels of noise. Thus, the SVM model might not have been able to create a classifier that could be sufficiently general for noisy future data, due to hindered class representation.

~

\noindent\textbf{3) Can denoising methods help in these situations?}

Overall, the use of denoising methods improved the classification performance when both training and test sets were affected by the same type of noise. However, the achieved result was not as good as the one obtained using the original dataset, probably due to the loss of detail and texture caused by these methods. Note, however, that models created with images after denoising did not perform well when tested with noisy images.

\subsection{Supplementary material}

Due to the size restrictions, not all results were presented in this paper. These results are available at: \iffinal
    \href{https://github.com/gbpcosta/wvc\_2016\_noise}{https://github.com/gbpcosta/wvc\_2016\_noise}.
\else
    Due to anonymity purposes this url was omitted.
\fi

\section{Conclusion}
\label{sec:conclusion}

Results presented in the previous section show that test classifiers in images with a different type of noise not only confuses the models, but also causes the problem become harder. This is noticeable on the diagonal of each heatmap, where none of the classifiers were able to overcome the performance of the classifier trained and tested with the original dataset.

When denoising is applied, the results obtained by classifying images from the same category (same type of noise or denoising method) were slightly better then the ones achieved by classifying noisy images. However, due to the smoothing caused by these methods, these results did not match the classification performance of the original dataset.

Future work include the analysis of the effect of noise in video descriptors, since temporal information might help overcome the difficulty of describing noisy data. The analysis performed in this paper should also be extended to include more datasets, descriptors and denoising methods, mainly to include deep learning methods, since these represent the state-of-the-art of image classification. Finally, the use of image quality metrics such as PSNR and SSIM can be important on comparing degraded images.




\iffinal
\section*{Acknowledgment}

This work was supported by FAPESP (grants \#2014/21888-2, \#2015/04883-0 and 	\#2015/05310-3).
\fi



\bibliography{ref}
%



\end{document}